\documentclass[fleqn,10pt]{wlscirep}
\usepackage[utf8]{inputenc}
\usepackage[T1]{fontenc}
\usepackage[left]{lineno}
\usepackage[misc]{ifsym}
\DeclareUnicodeCharacter{202F}{ }

\title{Urban Flood Observations: A global high-resolution hand-labeled dataset of post-flood inundation}

\author[1, 2, $\dag$, \Letter]{Rohit Mukherjee}
\author[2,$\dag$]{Hannah K. Friedrich}
\author[2, 7, \Letter]{Beth Tellman}
\author[2]{Ariful Islam}
\author[2, 6]{Zhijie Zhang}
\author[2, 5]{Jonathan Giezendanner}
\author[3]{Upmanu Lall}
\author[4]{Venkataraman Lakshmi}

\affil[1]{Pacific Northwest National Laboratory, Richland, WA, USA}
\affil[2]{University of Arizona, Tucson, USA}
\affil[3]{Columbia University, New York, USA}
\affil[4]{University of Virginia, Charlottesville, USA}
\affil[5]{Massachusetts Institute of Technology, Cambridge, Massachusetts, USA}
\affil[6]{Utah State University, Logan, Utah, USA}
\affil[7]{University of Wisconsin--Madison, Madison, WI, USA}
\affil[$\dag$]{these authors contributed equally to this work}
\affil[ \Letter]{email:
    \href{mailto:rohit.mukherjee@pnnl.gov}{rohit.mukherjee@pnnl.gov},
  \href{mailto:tellman2@wisc.edu}{tellman2@wisc.edu}
}

\begin{abstract}
  Urban flooding affects lives and infrastructure worldwide. Mapping post-flood inundation in complex urban environments from satellite imagery remains challenging due to limited spatial resolution, infrequent acquisitions, and cloud cover. We present Urban Flood Observations (UFO),
  a global, hand-labeled dataset of post-flood inundation in diverse urban settings. UFO comprises 215 image chips (1024$\times$1024 pixels) from 14 flood events between 2017 and 2021, derived from 3\,m PlanetScope imagery. Each chip is annotated with two
  classes: ``inundated'', covering all visible surface water (both floodwater and pre-existing permanent or seasonal water bodies), and ``non-inundated''. To demonstrate the dataset's utility, we trained a segmentation model using leave-one-event-out cross-validation, achieving a mean Intersection over
  Union (IoU) of 77.3\%. We also used UFO to evaluate two widely used surface water products, the Sentinel-1-based NASA IMPACT model and Google's 10\,m Dynamic World water and flooded-vegetation classes, which yielded IoUs of 44.1\% and 48.1\%, respectively. UFO is publicly available to support the development and validation of urban inundation mapping methods.
\end{abstract}

\begin{document}


\flushbottom
\maketitle

\thispagestyle{empty}


\section*{Background \& Summary}

Urban flooding poses a global risk, affecting billions of people and damaging critical infrastructure, resulting in substantial economic losses and loss of life~\cite{hammond2015}. The severity of this threat is increasing due to climate change, which drives more extreme rainfall events~\cite{ren_rising_2023}, and rapid urbanization, which heightens flood risks in densely populated areas~\cite{zhang2018b, zhou_comparison_2019, tellman_satellite_2021, dottori_2018_increased, rentschler_flood_2022}. Accurate mapping and monitoring of inundation extents are therefore critical for effective disaster response, infrastructure resilience, and risk mitigation in urban settings. Satellite remote sensing provides a scalable means of addressing this need, yet mapping urban inundation from space remains an open challenge.

Synthetic aperture radar (SAR)~\cite{amitrano2024flood} and multispectral optical sensors have been applied to inundation mapping~\cite{munawar2022remote, j-p_schumann_breakthroughs_2024}. However, both modalities face limitations in dense urban areas. SAR-based approaches, including those using Sentinel-1, show reduced flood detection accuracy in built-up environments due to complex backscattering interactions with urban structures~\cite{mason2021, amitrano2024flood}. Optical sensors, meanwhile, are limited by cloud cover during and after flood events~\cite{tellman_satellite_2021} and by the coarser spatial resolution of publicly available products, which often cannot resolve inundated pixels in dense urban areas~\cite{bentivoglio2022a}. Commercial satellite constellations like PlanetScope (PS; optical) and ICEYE (radar) partially address the resolution and revisit limitations through higher spatial resolution and more frequent acquisitions~\cite{tarpanelli2026potential, olthof2020testing, ardila2022persistent}, though their data are less accessible than public alternatives.

Deep learning methods have shown promise for segmenting inundation in satellite imagery~\cite{bentivoglio2022a, giezendanner2023b, peng2019, zhao2022}, yet their performance depends heavily on the availability of well-annotated training data representative of target flood conditions~\cite{bonafilia2020b}. Several flood-related datasets have been curated for machine learning. At medium resolution, Sen1Floods11~\cite{bonafilia2020b}, Kuro Siwo~\cite{bountos2023kuro}, STURM-Flood~\cite{notarangelo2025sturm}, SenForFlood~\cite{matosak2025senforflood}, and UrbanSARFloods~\cite{zhao2024} provide Sentinel-based flood labels, while WorldFloods~\cite{mateo-garcia2021d} is a dedicated optical-only benchmark. Multimodal datasets like MMFlood~\cite{montello2022}, S1S2-Water~\cite{wieland2024b}, and Sen2GF3Floods~\cite{chen2026sen2gf3floods} support multisensor flood mapping by fusing optical and SAR imagery. NASA's manually labeled inundation dataset~\cite{gahlot2021flood} also operates at medium resolution. Although UrbanSARFloods specifically targets urban flooding, its 10\,m resolution can be insufficient for capturing small-scale features in complex built environments~\cite{bentivoglio2022a}. At higher resolution, SpaceNet-8~\cite{hansch2022a}, FloodNet~\cite{rahnemoonfar2021}, GF-FloodNet~\cite{zhang2023gffloodnet}, and DeepFlood~\cite{fawakherji2025} provide sub-meter to meter-scale labels from aerial, UAV, or high-resolution satellite imagery, but remain limited in geographic scope, event diversity, or urban specificity. FloodPlanet~\cite{zhang2025floodplanet} provides hand-labeled chips from 3\,m PlanetScope imagery across 19 global events, demonstrating that high-resolution labels improve inundation detection even when applied to lower-resolution Sentinel-1 and Sentinel-2 inputs; however, it does not specifically target urban environments.

Building on these efforts, we introduce the \emph{Urban Flood Observations (UFO)} dataset~\cite{mukherjee_2025_ufo}: a global, hand-labeled collection of post-flood PlanetScope imagery centered on urban inundation. UFO contains 215 chips (1024$\times$1024 pixels) at 3\,m resolution, spanning 14 flood events between 2017 and 2021 across pluvial, fluvial, and storm-surge drivers. The ``inundated'' class is defined as all visible surface water in post-flood imagery, including floodwater and pre-existing water bodies (permanent or seasonal). A SegFormer-B2 model trained on UFO under leave-one-event-out cross-validation reaches a mean Intersection over Union (IoU) of 77.3\% across 14 held-out events. Using UFO as a 3\,m reference, two widely used 10\,m products, the NASA IMPACT Sentinel-1 model~\cite{paul2021} and Google's Dynamic World water and flooded-vegetation classes~\cite{brown2022b}, reach IoUs of 44.1\% and 48.1\%, respectively. UFO supports the development and validation of urban inundation mapping methods at 3\,m resolution.

\section*{Methods}

The preparation of the UFO dataset involved four steps: (1)~identifying and selecting urban flood events, (2)~downloading and processing PlanetScope imagery, (3)~annotating inundation extents, and (4)~performing quality control on the labeled data. Throughout this paper we use the term \emph{inundation} to describe all visible surface water during and after a flood event, as this is what remote sensing data can detect. The term \emph{flood} is used more broadly to describe events whose social or economic consequences are damaging to lives, livelihoods, or property.

\subsection*{Event Selection and Documentation}

\begin{figure}[!htbp]
  \centering
  \includegraphics[width=\linewidth]{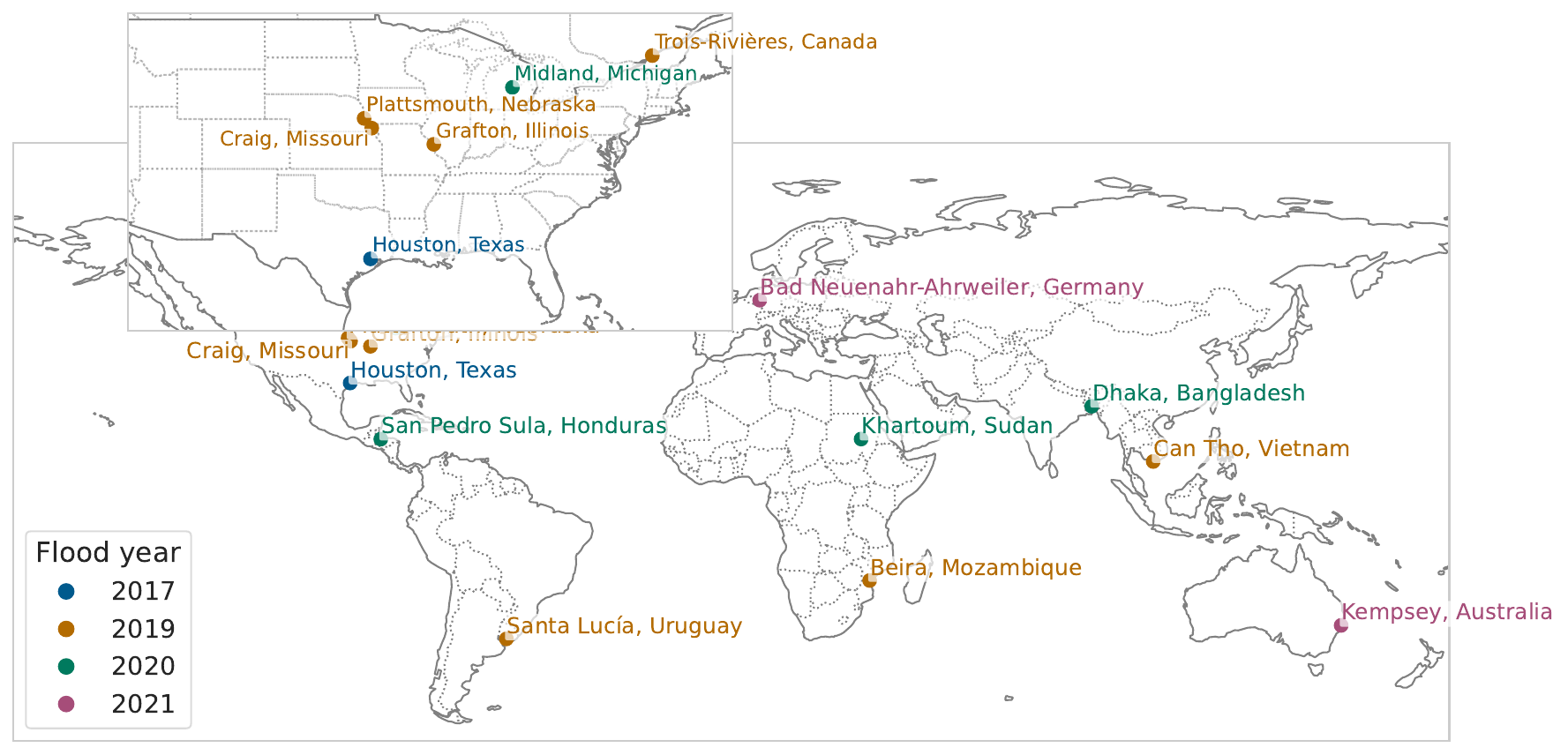}
  \caption{Locations of the 14 urban flood events in the UFO dataset~\cite{mukherjee_2025_ufo}.}
  \label{fig:globalEvent}
\end{figure}

To identify candidate urban flood events, we curated a list from news coverage on FloodList~\cite{floodlist_2024} (floodlist.com), restricting the search to events from 2017 onward, when consistent PlanetScope coverage became available. We selected events to span a range of geographic regions (six continents) and urban morphologies (from dense informal settlements to low-density suburban areas). The events also cover three flood-driver types: most are fluvial (river overflow), while Hurricane Harvey (Houston) is predominantly pluvial and Cyclone Idai (Beira) is a storm-surge–driven coastal event. Table~\ref{tab:eventsSummary} summarizes the 14 selected events and Figure~\ref{fig:globalEvent} shows their global distribution.

\begin{table}[!htbp]
  \centering
  \scriptsize
  \resizebox{1\textwidth}{!}{
    \begin{tabular}{lllll}
      \toprule
      \textbf{SiteID} & \textbf{Location} & \textbf{Flood cause and estimated dates} & \textbf{PlanetScope dates} & \textbf{Flood driver}\\
      \midrule
      BNA & Bad Neuenahr-Ahrweiler, Germany & Extreme rainfall, 13–16 Jul 2021 & 18 Jul 2021 & Fluvial\\
      BEI & Beira, Mozambique & Cyclone Idai, 14–16 Mar 2019 & 25 Mar 2019, 26 Mar 2019 & Storm surge\\
      CMO & Craig, Missouri, USA & Missouri R. floods, 10 Jan–29 Jul 2019 & 21 Mar 2019 & Fluvial\\
      CTO & Can Tho, Vietnam & High water levels + tide, 30 Sep–1 Oct 2019 & 29 Sep 2019, 2 Oct 2019 & Fluvial\\
      DKA & Dhaka, Bangladesh & Major riverine flooding, Jun–Oct 2020 & 25 Aug 2020 & Fluvial\\
      GIL & Grafton, Illinois, USA & Mississippi R. floods, 10 Jan–29 Jul 2019 & 4 May 2019, 7 May 2019, 14 May 2019 & Fluvial\\
      HTX & Houston, Texas, USA & Hurricane Harvey, 25–30 Aug 2017 & 31 Aug 2017 & Pluvial\\
      KTM & Khartoum, Sudan & Heavy rainfall, mid-Jul–Sep 2020 & 29 Aug 2020 & Fluvial\\
      MID & Midland, Michigan, USA & Dam burst, 19 May 2020 & 20 May 2020 & Fluvial\\
      NSW & Kempsey, NSW, Australia & Heavy rainfall, 17–26 Mar 2021 & 23 Mar 2021 & Fluvial\\
      PNE & Plattsmouth, Nebraska, USA & Missouri R. floods, 10 Jan–29 Jul 2019 & 18 Mar 2019, 27 Mar 2019 & Fluvial\\
      QUE & Trois-Rivières, Quebec, Canada & Snowmelt, Apr–Jun 2019 & 29 Apr 2019 & Fluvial\\
      SLC & Santa Lucía, Uruguay & Heavy rain, 11–15 Jun 2019 & 20 Jun 2019 & Fluvial\\
      SPS & San Pedro Sula, Honduras & Hurricanes Eta \& Iota, 3–18 Nov 2020 & 25 Nov 2020 & Fluvial\\
      \bottomrule
  \end{tabular}}
  \caption{\label{tab:eventsSummary}The 14 flood events in the UFO dataset: location, flood cause and estimated dates, PlanetScope image acquisition dates, and flood drivers. SiteID codes (BNA, BEI, etc.) are used in subsequent figures.}
\end{table}

For each event we documented its approximate start and end dates, the urban area sampled, and the dates of PlanetScope acquisitions. High-profile events such as Hurricane Harvey had well-reported timelines; for events with prolonged or recurring flooding, such as the 2019 Mississippi River floods, we estimated temporal bounds from available news reports.

We then queried PlanetScope imagery using the Planet Explorer tool (planet.com/explorer/). A scene was retained only if it had ${<}30$\% cloud cover and contained visible inundation. Many candidate events were excluded because imagery was either unavailable during flooding or cloud-free only after floodwater had receded. Each scene intersecting the approximate flood footprint was visually inspected for cloud obstruction and inundation visibility before inclusion.

\subsection*{PlanetScope Image Processing}

For each retained event, we downloaded PlanetScope four-band surface reflectance imagery (blue, green, red, near-infrared). Each scene was subdivided into 1024$\times$1024-pixel chips ($3.072\times3.072$\,km at 3\,m resolution). To aid labelers in identifying surface water, we generated three composites per chip: (1)~true-color (red, green, blue), (2)~false-color (near-infrared, red, green), and (3)~a normalized difference water index (NDWI) map, computed using the green and near-infrared bands.

Within each retained scene, chips containing a mix of built-up features and visible post-flood inundation were selected through visual inspection of the imagery. The selection prioritized areas of maximum observable inundation within complex urban landscapes; dense built-up zones with heterogeneous land cover were preferred over homogeneous open areas. No minimum water fraction threshold was applied; rather, chips were chosen qualitatively based on evidence of post-flood inundation in the imagery. The selection was guided by the goal of assembling a challenging validation dataset representative of real-world urban flood conditions, rather than by considerations of deep learning training balance (e.g., class ratio or scene diversity targets). Chips with no discernible post-flood inundation were excluded, and only chips with predominantly clear skies were retained, though some included minor partial cloud cover. This targeted, non-random selection means the dataset is not a spatially exhaustive map of any single flood event; rather, it samples the most clearly inundated built-up areas across events. Users should be aware of this sampling strategy when interpreting dataset-wide statistics such as aggregate inundation fraction.

\subsection*{Data Labeling}

\subsubsection*{Class Definitions}

UFO uses a binary labeling scheme. The inundated class encompasses all pixels where surface water is clearly visible, including floodwater and pre-existing water bodies (permanent or seasonal). The non-inundated class covers all remaining pixels, which may include urban surfaces, vegetation, bare soil, or clouds. Only fully inundated pixels were labeled; pixels representing flooded vegetation (i.e., areas where water may be present beneath a vegetation canopy) were classified as non-inundated.

This ontology was chosen for three reasons. First, floodwater and pre-existing water bodies (permanent or seasonal) are visually indistinguishable in single-date post-flood PlanetScope imagery. Second, distinguishing floodwater from pre-existing water in a single image is inherently ambiguous: it depends on temporal frequency rather than spectral appearance, and existing global surface water products disagree substantially~\cite{rajib_call_2024, venter_2022_global}, in part because no product operates at the 3--5\,m resolution of PlanetScope. Generating a pre-flood baseline mask is similarly difficult given seasonal variability in water body extent. Third, this inclusive definition aligns with the operational goal of machine learning segmentation: an effective flood detection algorithm must learn to recognize the full diversity of water appearances, from clear to heavily sedimented, rather than only the subset attributable to flooding. Users who require flood-only extent can difference UFO labels against surface water baselines derived from pre-flood imagery or ancillary water products that map pre-existing water, such as JRC Global Surface Water~\cite{pekel2016} or ESA WorldCover~\cite{zanaga_esa_2021}, depending on the use case.

\subsubsection*{Annotation Process}

Labeling was performed on the Labelbox platform~\cite{labelbox_online_2025} (labelbox.com). Labelers were provided with the true-color, false-color, and NDWI composites for each chip. Using these composites, they manually delineated all clearly visible surface water as inundated. All remaining pixels were labeled as non-inundated, yielding a binary mask per chip.

\begin{figure}[!htbp]
  \centering
  \includegraphics[width=\linewidth]{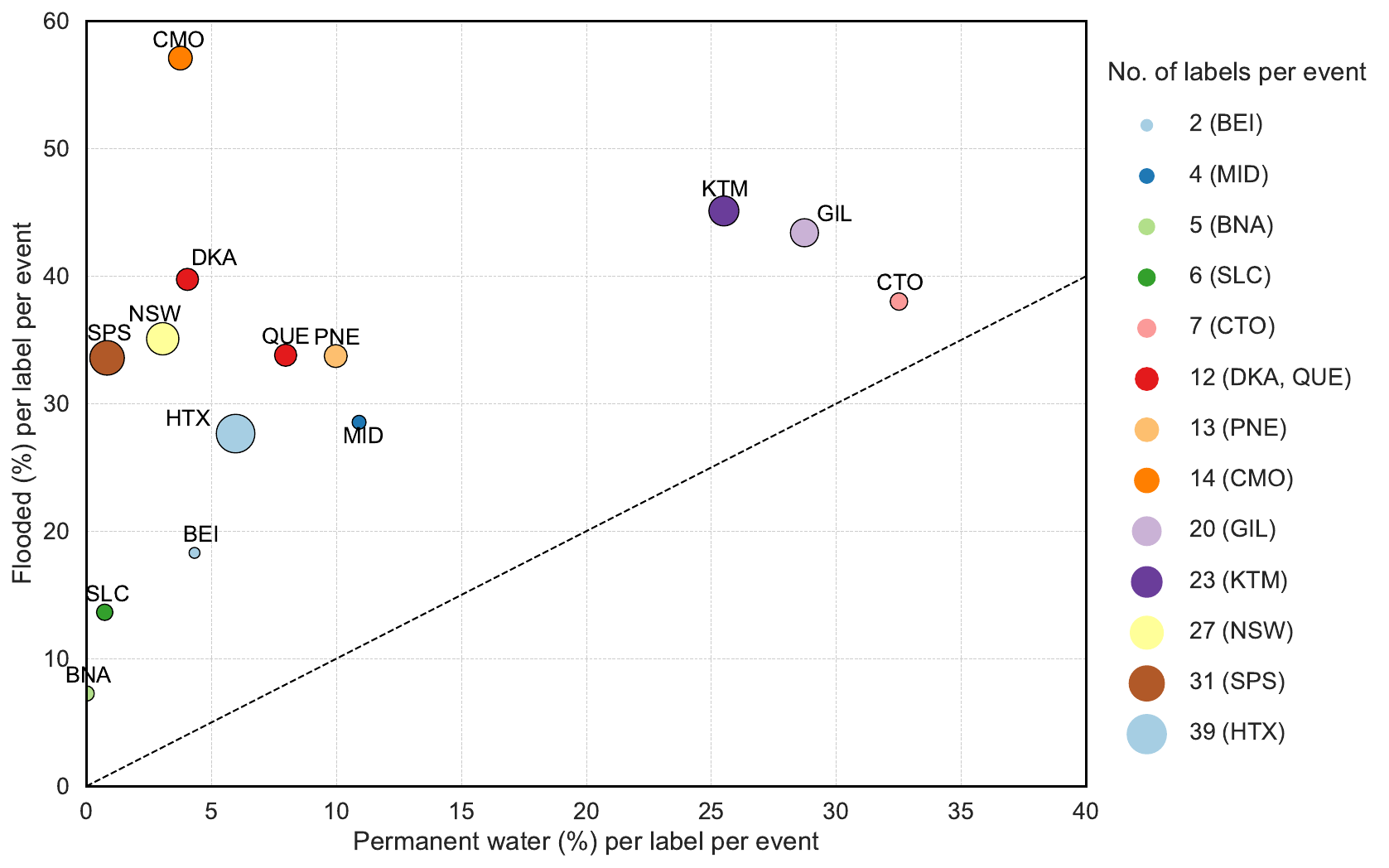}
  \caption{Per-event flooded area vs.\ permanent water area, expressed as percentages of total chip area. UFO's ``inundated'' class encompasses all visible surface water; ``flooded \%'' is computed as inundated area minus permanent water estimated from ESA WorldCover 2020~\cite{zanaga_esa_2021}. Point size indicates the number of labeled chips per event.}
  \label{fig:labelDist}
\end{figure}

\begin{figure}[!htbp]
  \centering
  \includegraphics[width=0.72\linewidth]{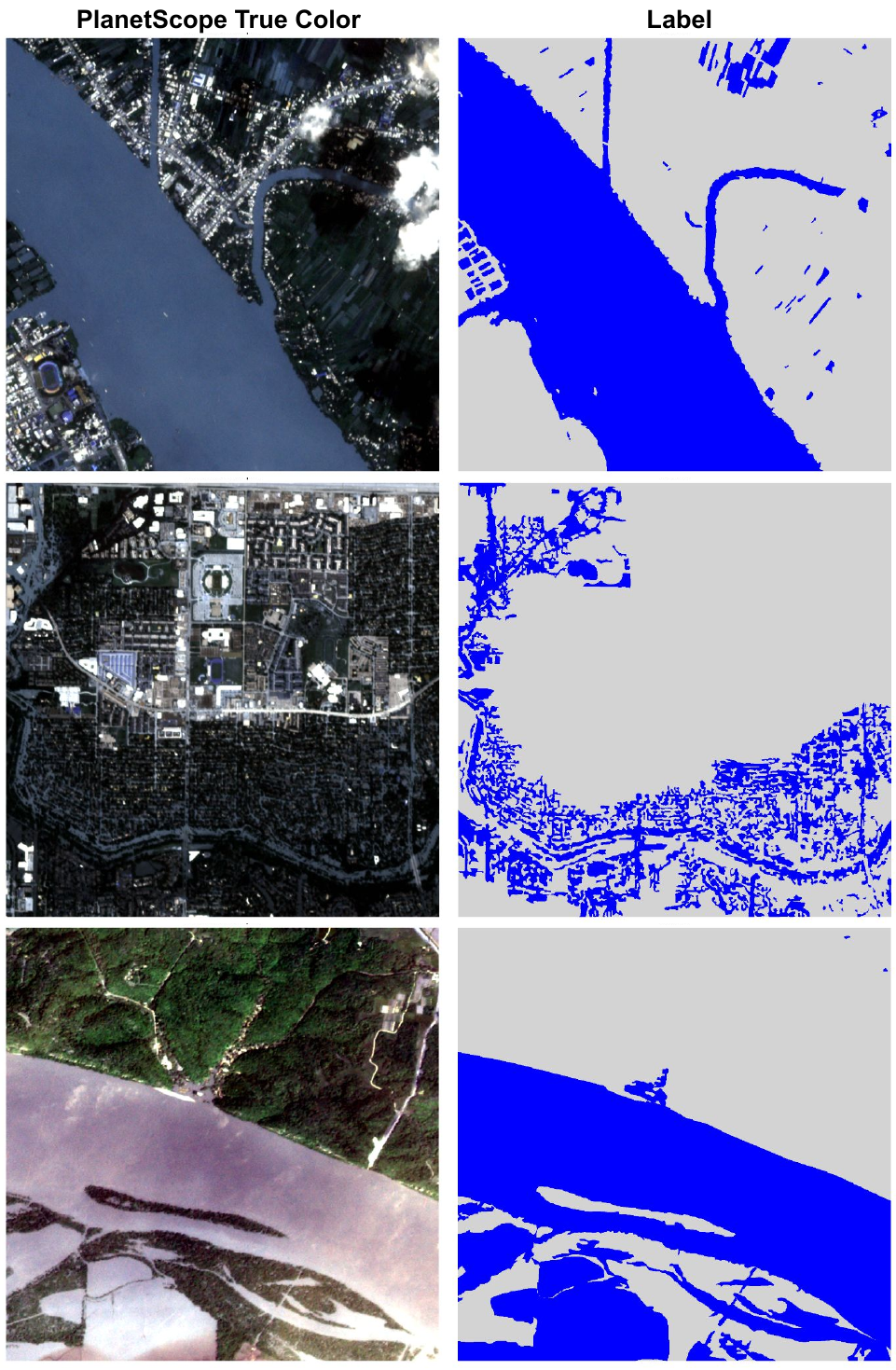}
  \caption{Examples of PlanetScope true-color chips (left) and corresponding post-flood inundation labels (right). Top: Can Tho, Vietnam (2019); middle: Houston, Texas (Hurricane Harvey, 2017); bottom: Grafton, Illinois (2019).}
  \label{fig:labelExamples}
\end{figure}

Figure~\ref{fig:labelDist} shows the distribution of inundated versus permanent water pixels per event, expressed as percentages of total chip area. Permanent water was estimated using the ESA WorldCover 2020 water class~\cite{zanaga_esa_2021}. The Craig (Missouri, USA) event had the highest inundation fraction per label, while Bad Neuenahr-Ahrweiler (Germany) and Can Tho (Vietnam) showed relatively lower fractions. Figure~\ref{fig:labelExamples} provides examples of PlanetScope imagery and their corresponding labels.

\subsection*{Quality Control}

Initial annotation was distributed across four research assistants with backgrounds in geospatial science and satellite-image interpretation; each chip was assigned to a single labeler (i.e., there was no overlapping annotation of the same chip by multiple labelers). After annotation, labels were exported and georeferenced to the UTM projection of their source PlanetScope scenes. Two successive quality-control (QC) rounds were then applied. In the first round, the labeling team collectively reviewed all chips, flagging mislabeled or ambiguous regions for re-labeling. A team reporting system captured difficult cases, including partially submerged vegetation, building shadows, and spectrally ambiguous surfaces. Chips with unresolvable ambiguities were discarded. In the second round, every remaining chip was reviewed, labels were revised where necessary, and the final set of chips for dataset inclusion was determined. This final pass ensured consistent labeling decisions across all events. Because no chips were independently annotated by multiple labelers, formal inter-annotator agreement metrics could not be computed; label reliability instead rests on the two-round QC process and a final single-reviewer pass over all labels.

Accurately delineating inundation at 3\,m resolution was challenging in heterogeneous urban landscapes where mixed land cover types (vegetation appearing bright red in false-color, urban surfaces appearing bright white in true-color, and transitional zones) coexisted in close proximity. Small clusters or individual pixels indicative of inundation (appearing teal in false-color composites) were sometimes missed or misclassified within these heterogeneous surroundings. Ambiguities also arose from partially submerged vegetation, which complicated the boundary between inundated and non-inundated areas (Figure~\ref{fig:labelExamples}, bottom row). In all such cases, only clearly visible surface water was labeled as inundated. Existing products such as Dynamic World~\cite{brown2022b} and Esri Land Cover~\cite{Karra2021} address this limitation by including flooded vegetation as a separate class. Variable illumination geometry, atmospheric scattering, and PlanetScope's sensor-to-sensor radiometric variability~\cite{frazier2021a} added further noise to visual interpretation. To mitigate these effects, the NDWI composite was used to suppress non-water pixels and highlight candidate water areas, and higher-resolution imagery from the Bing aerial basemap (bing.com/maps) was consulted to disambiguate features under non-flood conditions.

Following quality control, the final dataset contains 215 reliable inundation labels. During the review process, close to 100 labels were discarded; no chip was included without passing both QC rounds and final review.

\section*{Data Records}

The UFO dataset is publicly available on Zenodo~\cite{mukherjee_2025_ufo}. The dataset includes 215 pairs of annotated labels and corresponding PS four-band image chips (blue, green, red, near-infrared). All label images and their corresponding PS chips are 1024$\times$1024 pixels in size at 3\,m spatial resolution. It should be noted that, consistent with Planet Labs' data sharing policy for commercial imagery, the provided PS image chips are modified 8-bit representations of the original surface reflectance data product.

The shared dataset consists of an `UrbanFloodObservations' folder that contains the images and labels, and a `stac\_catalog' folder that holds the SpatioTemporal Asset Catalog (STAC) of the dataset. The `UrbanFloodObservations' folder contains two sub-folders, the names of which indicate the content: PS imagery and hand labels (labels). In the labels, the value 0 represents ``non-inundated'' pixels and the value 1 represents ``inundated'' pixels. The STAC catalog was created following the STAC specification for better interoperability, discoverability, and extensibility. It contains detailed data descriptions, including sensor band information, acquisition date, spatial resolution, and area covered, in the metadata. The `stac\_catalog' folder has the same structure as the `UrbanFloodObservations' folder, but stores the metadata as JSON files.

\section*{Technical Validation}

The UFO dataset is intended primarily as training and validation data for deep learning-based inundation segmentation. Below, we first confirm that the hand labels are learnable and transfer across unseen events, and then use UFO as a high-resolution reference to compare two widely used surface water products against the labels.

\subsection*{Learnability and Spatial Transferability}

To confirm that the UFO labels are learnable and transfer across unseen geographies, we trained a SegFormer-B2 model~\cite{xie2021segformer} adapted for four-band multispectral imagery (blue, green, red, and near-infrared). We selected SegFormer-B2 as a representative modern semantic segmentation architecture with a hierarchical encoder and lightweight decoder that efficiently aggregates multiscale context, and that has shown strong performance in remote sensing inundation mapping applications~\cite{jiang2024cropland, shaheen2025swinsegformer}. We initialized the network from ImageNet-pretrained weights and modified the first patch-embedding layer to accept four input channels. The model was trained on 256$\times$256 image tiles with a 10\% validation split within each training fold. Training used mixed precision for 30 epochs with a batch size of 8, a maximum learning rate of $6\times10^{-5}$, and weight decay of 0.01 under a one-cycle schedule. Data augmentation included random rotations, flips, dihedral transformations, brightness and contrast perturbations, and normalization. To address class imbalance while improving mask overlap, we optimized a combined Focal Loss and Dice Loss. Across all 215 chips, inundated pixels account for approximately 35.4\% of the total labeled area ($\sim$719\,km$^2$ out of $\sim$2{,}029\,km$^2$), underscoring the class imbalance motivating this loss design. During inference, we additionally applied test-time augmentation using multiple image scales and flip-based ensembling. Training and inference were performed on a single NVIDIA RTX 3080 (10\,GB VRAM).

\subsubsection*{Evaluation Protocol}
Model performance was evaluated via pixel-level comparison between predicted inundation maps and the UFO hand-labels. A confusion matrix was generated for each prediction, quantifying true positives (TP), false positives (FP), true negatives (TN), and false negatives (FN). From these counts, we computed Precision, Recall (Sensitivity), Specificity, F1 score, Overall Accuracy, and IoU. We report the mean and standard deviation for each metric across events to capture both central tendency and variability.

\subsubsection*{Cross-Validation Design and Results}

We employed a leave-one-event-out cross-validation strategy to test whether models trained on UFO generalize to flood events not seen during training. In each of 14 iterations, 13 events served as training data and the remaining event was held out entirely as an unseen test set. Consistent hyperparameters were maintained across all folds.

\begin{figure}[!htbp]
  \centering
  \includegraphics[width=0.95\linewidth]{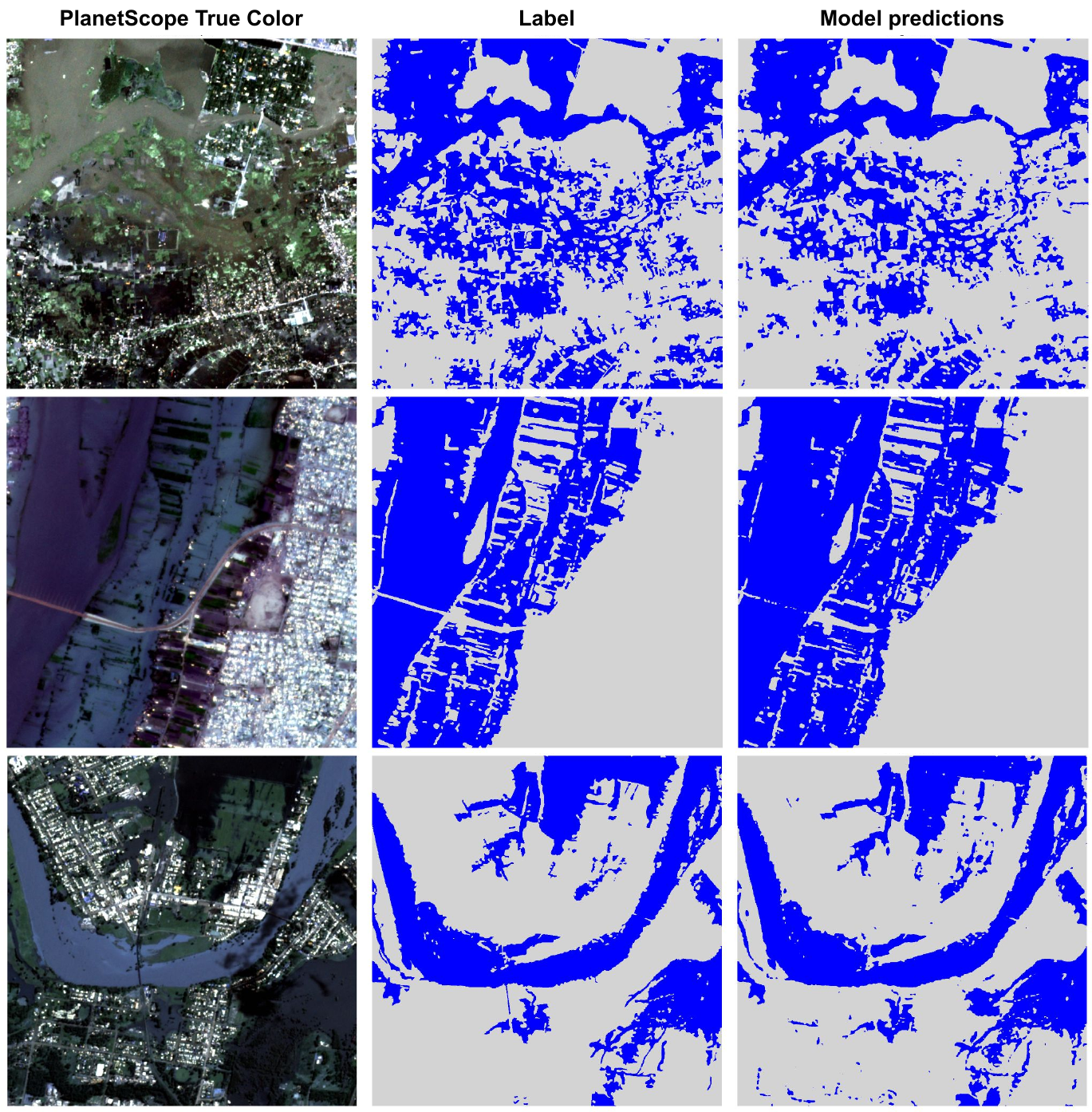}
  \caption{Predictions from the UFO-trained SegFormer-B2 model~\cite{xie2021segformer}. Columns: PlanetScope true-color image (left), UFO hand label (center), model prediction (right). Top to bottom: Dhaka, Bangladesh (2020); Khartoum, Sudan (2020); Kempsey, NSW, Australia (2021).}
  \label{fig:SegFormerRes}
\end{figure}

\begin{figure}[!htbp]
  \centering
  \includegraphics[width=0.85\linewidth]{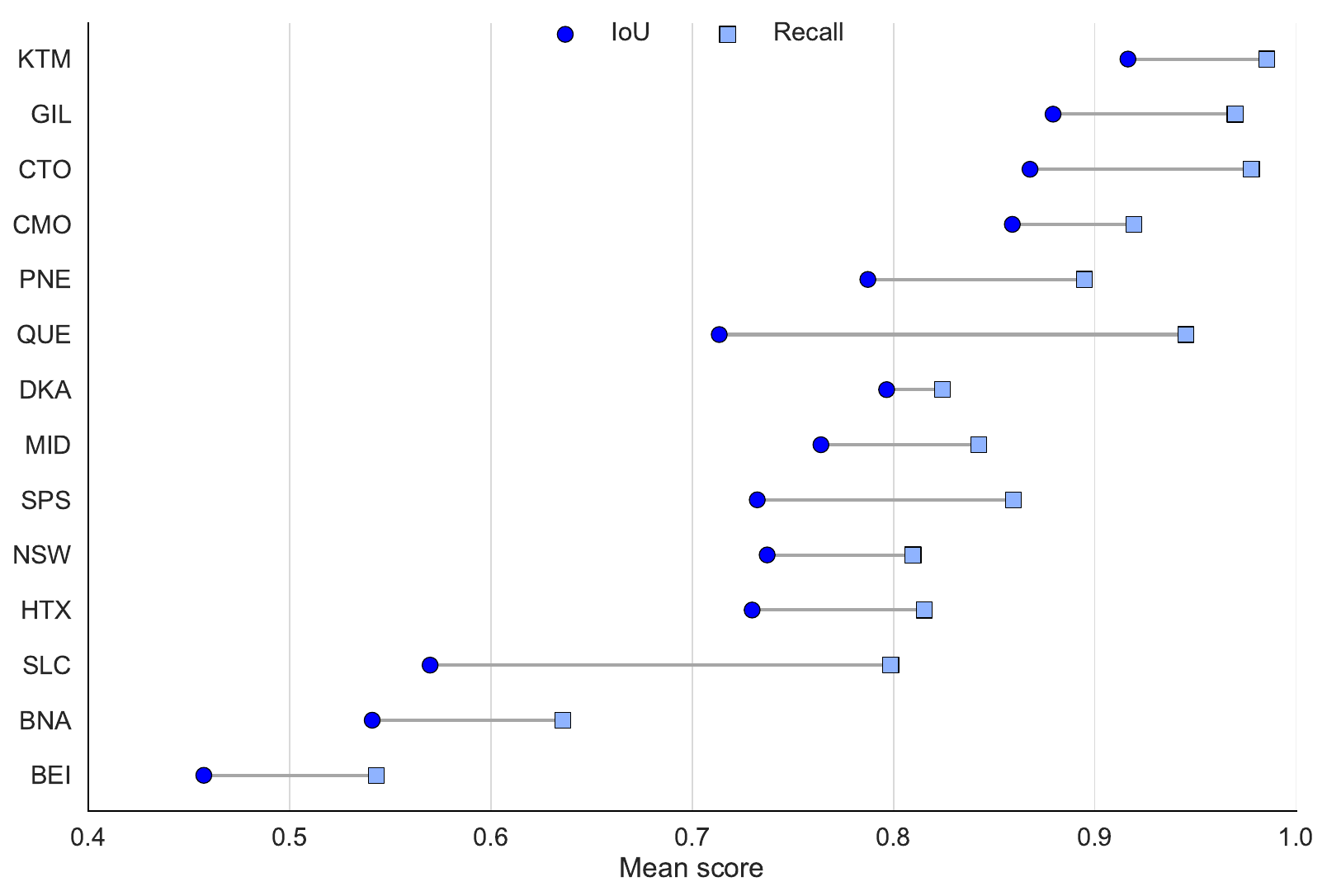}
  \caption{Per-event mean Intersection over Union (IoU) and Recall (Sensitivity) for the UFO-trained SegFormer-B2 under leave-one-event-out cross-validation. Error bars indicate one standard deviation across chips within each event.}
  \label{fig:evalIoURecall}
\end{figure}

Qualitative results (Figure~\ref{fig:SegFormerRes}) show that the model delineates coherent inundation extents across diverse urban landscapes. Quantitative performance is summarized in Table~\ref{table:performanceMetrics} and visualized per event in Figure~\ref{fig:evalIoURecall}. The model achieved a mean IoU of 77.3\% across all held-out events, indicating that the UFO labels are sufficient for a standard architecture to generalize across 14 geographically and climatologically distinct flood scenarios.

\begin{table}[!htbp]
  \centering
  \renewcommand{\arraystretch}{1.3}
  \begin{tabular}{|c|cc|cc|cc|cc|cc|cc|}
    \hline
    & \multicolumn{2}{c|}{\textbf{Precision}} & \multicolumn{2}{c|}{\textbf{Recall}} & \multicolumn{2}{c|}{\textbf{Specificity}} & \multicolumn{2}{c|}{\textbf{F1}} & \multicolumn{2}{c|}{\textbf{IoU}} & \multicolumn{2}{c|}{\textbf{Accuracy}} \\
    \textbf{Event} & mean & std & mean & std & mean & std & mean & std & mean & std & mean & std \\
    \hline
    BEI & 62.6 & 33.1 & 54.3 & 40.7 & 95.7 & 2.0  & 57.7 & 37.9 & 45.7 & 38.8 & 91.6 & 0.9 \\
    BNA & 77.7 & 5.7  & 63.6 & 12.4 & 98.6 & 0.5  & 69.7 & 9.3  & 54.1 & 11.4 & 95.9 & 1.7 \\
    CMO & 93.4 & 11.1 & 91.9 & 16.5 & 88.0 & 17.5 & 91.2 & 13.2 & 85.9 & 18.1 & 92.6 & 14.1 \\
    CTO & 88.5 & 8.4  & 97.8 & 2.0  & 91.6 & 7.6  & 92.7 & 5.0  & 86.8 & 8.5  & 94.3 & 4.3 \\
    DKA & 95.6 & 3.5  & 82.4 & 11.7 & 97.6 & 1.7  & 88.2 & 7.6  & 79.7 & 12.1 & 93.1 & 3.5 \\
    GIL & 90.4 & 17.5 & 97.0 & 1.7  & 94.7 & 6.7  & 92.4 & 14.3 & 87.9 & 17.0 & 95.5 & 4.4 \\
    HTX & 87.6 & 9.7  & 81.5 & 18.3 & 96.6 & 3.4  & 83.1 & 13.1 & 73.0 & 16.9 & 94.2 & 3.4 \\
    KTM & 93.0 & 8.3  & 98.5 & 1.6  & 94.3 & 6.4  & 95.4 & 4.9  & 91.6 & 8.1  & 97.1 & 1.8 \\
    MID & 88.6 & 9.3  & 84.2 & 14.5 & 96.3 & 2.5  & 85.9 & 10.5 & 76.4 & 16.5 & 94.6 & 2.3 \\
    NSW & 87.0 & 19.4 & 81.0 & 32.3 & 94.2 & 7.8  & 79.9 & 30.0 & 73.7 & 30.3 & 96.1 & 2.6 \\
    PNE & 85.6 & 11.0 & 89.5 & 15.3 & 88.8 & 15.6 & 87.0 & 12.0 & 78.7 & 17.1 & 94.4 & 6.3 \\
    QUE & 74.0 & 19.2 & 94.5 & 4.6  & 85.7 & 10.5 & 81.9 & 13.7 & 71.3 & 18.8 & 90.3 & 6.1 \\
    SLC & 68.2 & 15.3 & 79.9 & 13.5 & 94.1 & 3.6  & 71.8 & 11.3 & 57.0 & 13.1 & 91.6 & 4.5 \\
    SPS & 83.0 & 15.1 & 86.0 & 18.2 & 92.4 & 6.8  & 82.8 & 15.9 & 73.2 & 19.7 & 92.6 & 3.6 \\
    \hline
    \textbf{Overall} & 86.7 & 14.7 & 87.1 & 18.8 & 93.5 & 8.6  & 85.3 & 16.8 & 77.3 & 20.3 & 94.2 & 5.3 \\
    \hline
  \end{tabular}
  \caption{Leave-one-event-out cross-validation performance of SegFormer-B2 trained on UFO, reported per flood event (mean $\pm$ standard deviation across chips, in percentages). The \textbf{Overall} row aggregates statistics over all 215 chips.}
  \label{table:performanceMetrics}
\end{table}

Performance varied across events: IoU exceeded 86\% for Khartoum (KTM; 91.6\%), Grafton (GIL; 87.9\%), and Can Tho (CTO; 86.8\%), while it was lower for Beira (BEI; 45.7\%), Bad Neuenahr-Ahrweiler (BNA; 54.1\%), and Santa Lucía (SLC; 57.0\%). Several factors may contribute to this per-event variability, including differences in urban density, sediment load affecting water appearance, and the timing of image acquisition relative to peak inundation. We did not systematically test which of these factors drives the observed differences, and disentangling their contributions would require a larger and more balanced dataset than UFO currently provides.

\subsection*{Comparison with Existing Surface Water Products}

To assess whether UFO can function as a high-resolution reference for urban inundation mapping, we compared the hand labels against two widely used, independently generated medium-resolution products: (i) the NASA IMPACT-based flood mapping model (S1-IMPACT)~\cite{paul2021}, which uses 10\,m Sentinel-1 SAR data, and (ii) Google's Dynamic World (S2-DW)~\cite{brown2022b}, a global 10\,m land cover product derived from Sentinel-2 optical imagery, from which we extracted and combined the ``water'' and ``flooded vegetation'' classes. For context, we also report the performance of the UFO-trained SegFormer-B2 model on the same labels.

\begin{figure}[!htbp]
  \centering
  \includegraphics[width=\linewidth]{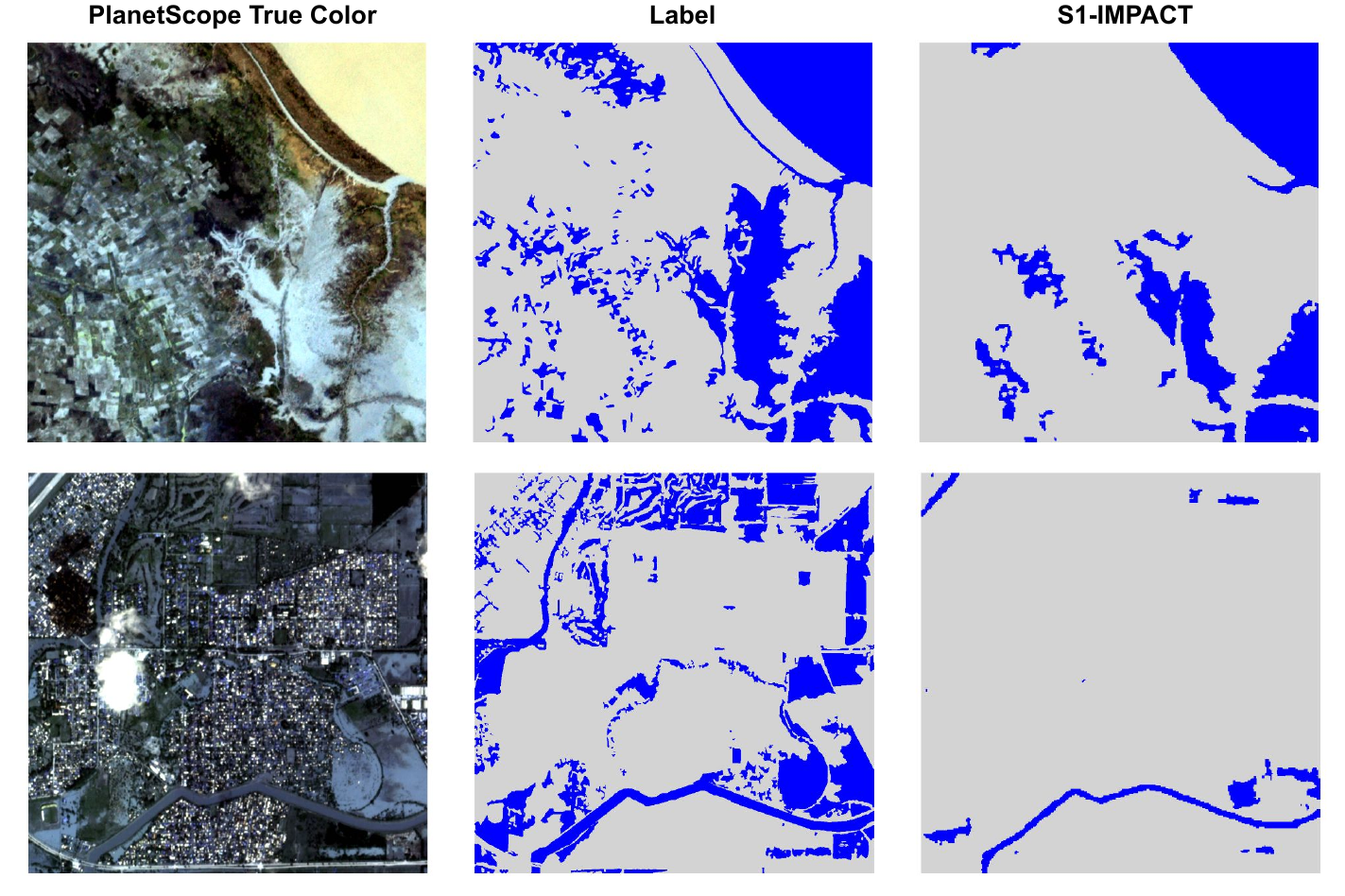}
  \caption{PlanetScope true-color imagery (left), UFO hand labels (center), and S1-IMPACT inundation predictions~\cite{paul2021} (right) for two same-day acquisitions. Top: Beira, Mozambique (2019); bottom: San Pedro Sula, Honduras (2020).}
  \label{fig:s1impactfig}
\end{figure}

\begin{figure}[!htbp]
  \centering
  \includegraphics[width=\linewidth]{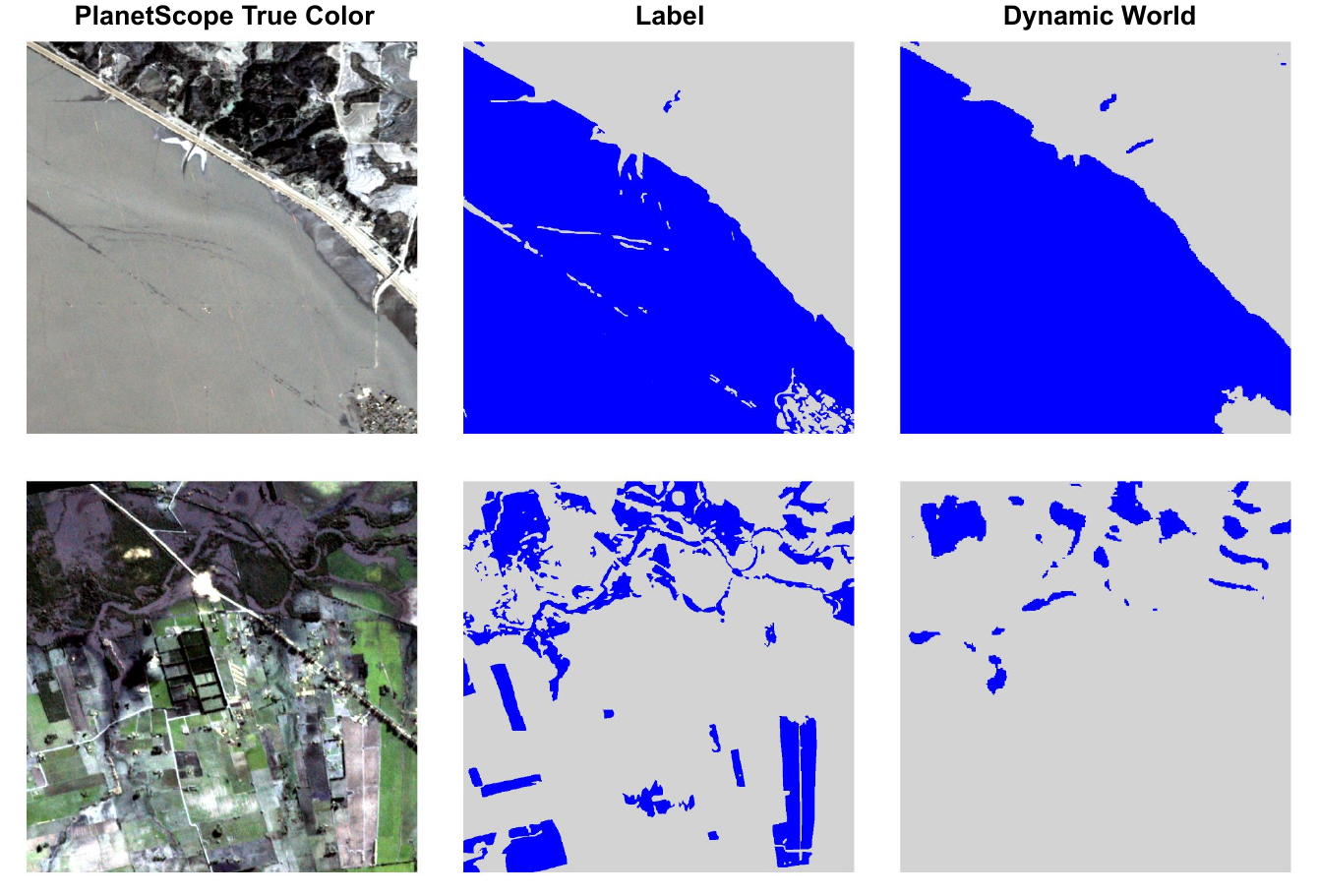}
  \caption{PlanetScope true-color imagery (left), UFO hand labels (center), and Dynamic World predictions~\cite{brown2022b} (right; water and flooded-vegetation classes combined) for two same-day acquisitions. Top: Craig, Missouri (2019); bottom: Santa Lucía, Uruguay (2019).}
  \label{fig:dwfigure}
\end{figure}

Comparison was restricted to temporally coincident (same day) acquisitions (and cloud-free for S2-DW), yielding 42 UFO labels for S1-IMPACT evaluation (Figure~\ref{fig:s1impactfig}) and 31 labels for S2-DW (Figure~\ref{fig:dwfigure}). Because the two products draw from different acquisition subsets, they were each evaluated independently against the UFO reference labels. Both 10\,m products were evaluated at 3\,m to match the native PlanetScope resolution.

\begin{figure}[!htbp]
  \centering
  \includegraphics[width=0.9\linewidth]{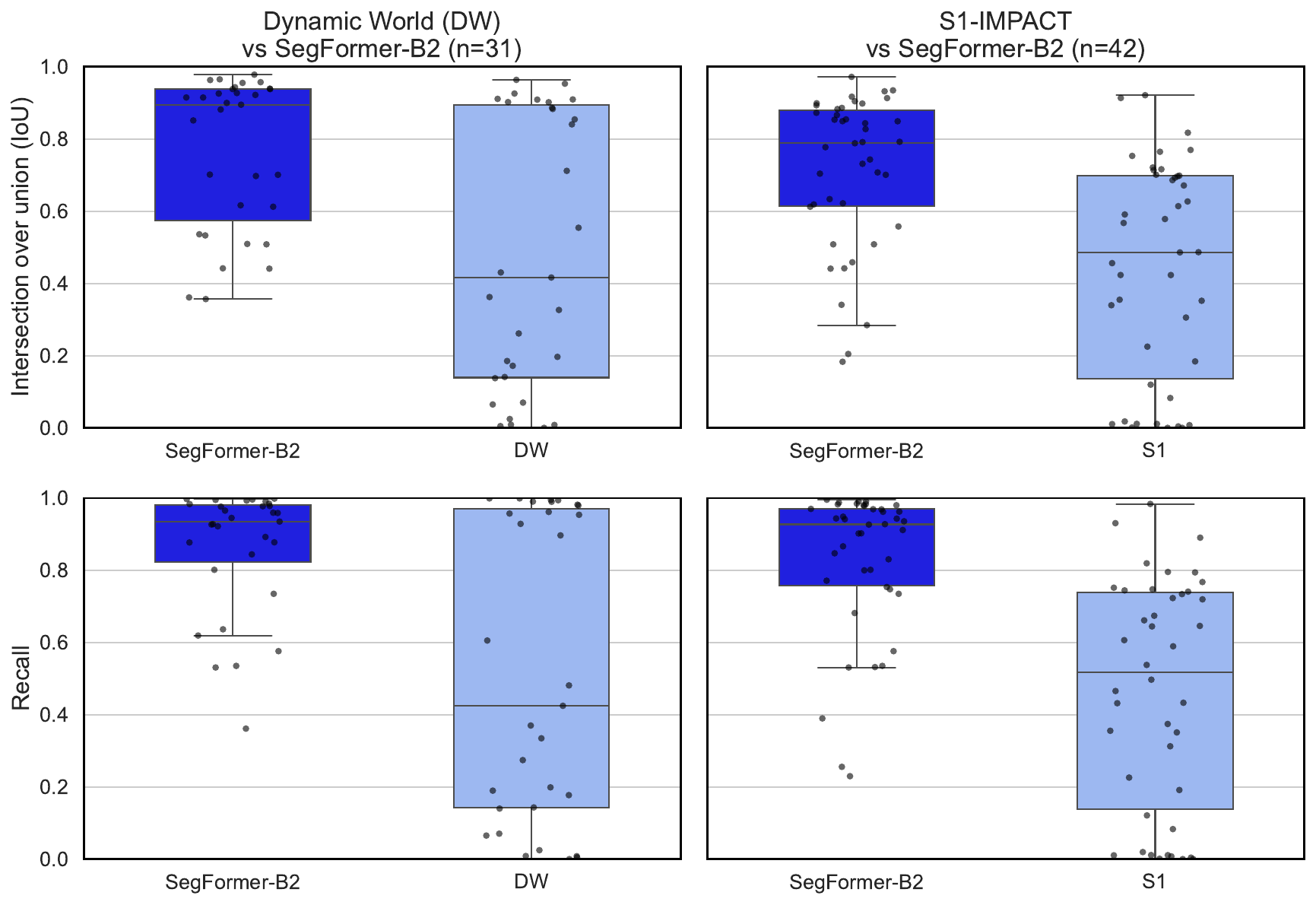}
  \caption{Chip-level Intersection over Union (IoU) and Recall (Sensitivity) against UFO labels. Grey boxplots show the UFO-trained SegFormer-B2; blue boxplots show benchmark products. Left: Dynamic World vs.\ UFO model ($n=31$). Right: S1-IMPACT vs.\ UFO model ($n=42$). Boxes show median and interquartile range; points denote individual chips. All comparisons use same-day acquisitions.}
  \label{fig:s1DWimpactVal}
\end{figure}

Table~\ref{table:performanceMetricsComparison} and Figure~\ref{fig:s1DWimpactVal} summarize the results. S1-IMPACT achieved high precision (mean: 86.2\%) but low recall (mean: 46.2\%), yielding an IoU of 44.1\%. S2-DW followed the same pattern: high precision (mean: 85.9\%) but low recall (mean: 52.1\%), resulting in an IoU of 48.1\%. Both products displayed substantial scene-to-scene variability. These results indicate that, relative to the 3\,m UFO labels, the two medium-resolution products identify many water pixels correctly but omit a substantial fraction of visible inundated area in the evaluated urban scenes. On the same image subsets, the UFO-trained SegFormer-B2 model reached 76.1\% IoU (vs.\ S1-IMPACT) and 82.2\% IoU (vs.\ S2-DW).

\begin{table}[!htbp]
  \centering
  \renewcommand{\arraystretch}{1.3}
  \begin{tabular}{|c|cc|cc|cc|cc|cc|cc|}
    \hline
    & \multicolumn{2}{c|}{\textbf{Precision}} & \multicolumn{2}{c|}{\textbf{Recall}} & \multicolumn{2}{c|}{\textbf{Specificity}} & \multicolumn{2}{c|}{\textbf{F1}} & \multicolumn{2}{c|}{\textbf{IoU}} & \multicolumn{2}{c|}{\textbf{Accuracy}} \\
    \textbf{Surface water models} & mean & std & mean & std & mean & std & mean & std & mean & std & mean & std \\
    \hline
    \textbf{S1-IMPACT} & 86.2 & 23.4 & 46.2 & 31.8 & 95.8 & 11.4 & 54.4 & 33.5 & 44.1 & 30.1 & 84.8 & 13.2 \\
    \textbf{UFO model}          & 82.0 & 16.0 & 91.2 & 6.6  & 94.6 & 4.1  & 85.4 & 11.5 & 76.1 & 15.6 & 94.7 & 2.3  \\
    \hline
    \textbf{S2-DW}     & 85.9 & 17.6 & 52.1 & 41.3 & 86.8 & 24.2 & 55.9 & 37.3 & 48.1 & 37.7 & 86.7 & 13.8 \\
    \textbf{UFO model}          & 86.4 & 14.3 & 93.4 & 7.2  & 94.6 & 6.9  & 89.3 & 10.5 & 82.2 & 16.2 & 96.5 & 2.5  \\
    \hline
  \end{tabular}
  \caption{Performance metrics (mean $\pm$ standard deviation across chips, in percentages) for two surface water products evaluated against UFO labels, alongside the UFO-trained SegFormer-B2 model (UFO model) on the same image subsets. Top: NASA IMPACT Sentinel-1 model (S1-IMPACT; $n=42$ chips). Bottom: Dynamic World Sentinel-2 water and flooded-vegetation classes (S2-DW; $n=31$ chips). All comparisons use same-day acquisitions; both 10\,m products were resampled to 3\,m to match PlanetScope.}
  \label{table:performanceMetricsComparison}
\end{table}

These comparisons should be interpreted with the differences among the datasets in mind. The 10\,m Sentinel-1/2 products cannot resolve some narrow or fragmented inundated areas visible in the 3\,m PlanetScope imagery, and the comparison subsets were limited to same-day acquisitions with suitable coverage. The benchmark is therefore intended as a technical check of UFO as a fine-resolution reference, rather than as a comprehensive assessment of the operational products.

\section*{Usage Notes}

UFO is suitable for supervised training of semantic segmentation models for post-flood inundation mapping. The leave-one-event-out cross-validation in the Technical Validation section provides a baseline for this use case. The 3\,m resolution captures fine-scale inundation features, including water in narrow streets and around buildings, that may not be resolved in publicly available 10\,m datasets. River width measurements after flooding are also likely to be more precise. To improve accuracy for flood events or regions not represented in UFO, a model trained on UFO can be fine-tuned with a small number of local labels to adapt to region-specific flood characteristics or urban morphologies. Predictions from UFO-trained models can also be used as pseudo-labels for new events or as an initial labeled product that can be further refined manually, ultimately reducing labeling effort.

Beyond model training, UFO can be used as a 3\,m reference for evaluating surface water and flood mapping products in urban settings. The same evaluation framework used in the Technical Validation can be applied to algorithms based on other satellite imagery products or outputs from hydrodynamic flood models. The accompanying STAC catalog includes acquisition dates and spatial metadata that facilitate identification of temporally coincident imagery from other sensors.

Because the UFO labeling ontology defines all visible surface water as inundated, including floodwater and pre-existing water bodies (permanent or seasonal), users requiring flood-only extent can difference UFO labels against water baselines, such as water masks from pre-flood satellite imagery products, JRC Global Surface Water~\cite{pekel2016} or ESA WorldCover~\cite{zanaga_esa_2021}. Figure~\ref{fig:labelDist} provides per-event estimates of permanent water fraction derived from WorldCover 2020. The georeferenced chips and STAC metadata also facilitate integration with complementary geospatial layers such as Digital Elevation Models, building footprints, or population density data for flood depth estimation, exposure analysis, or impact assessment.

Because the original PlanetScope surface-reflectance scenes cannot be redistributed under Planet Labs' commercial license, UFO provides the modified 8-bit chips together with the metadata required to recover the source imagery. Each chip and its label is georeferenced to the UTM projection of its source PlanetScope scene, and the accompanying STAC catalog records the acquisition date, spatial footprint, and band information for every chip. Users with PlanetScope data access can use this date-and-location metadata to retrieve the corresponding original PlanetScope scenes and align them with the UFO labels, enabling reuse of the annotations with the full-resolution 16-bit imagery where required.

Several characteristics of the data warrant attention when designing analyses. The distributed PlanetScope chips are uint8 (8-bit) representations of the original surface reflectance product, consistent with Planet Labs' data sharing policy; this reduces radiometric dynamic range, and models trained on these chips will require normalization adjustments if applied to workflows using the native 16-bit imagery. Chips were manually selected to capture clearly inundated built-up areas within each event, with a preference for dense urban zones over homogeneous open areas, rather than sampled randomly. The dataset is therefore not a spatially exhaustive map of any single flood footprint. Pixels where water is obscured by tree canopy or dense vegetation are labeled as non-inundated, so total flood extent may be underestimated in heavily vegetated areas relative to products that include a flooded-vegetation class~\cite{brown2022b, Karra2021}. The 14 events span diverse geographies and three flood-driver types (though most are fluvial) but cannot represent all possible urban flood conditions; events with very few chips (e.g., Beira with 2, Midland with 4) offer limited within-event variability. Finally, all imagery is cloud-dependent, so peak inundation occurring under cloud cover may not be captured, and each label corresponds to a single time-specific snapshot rather than a multi-temporal sequence.

\section*{Data availability}

The Urban Flood Observations (UFO) dataset (v2), including PlanetScope image chips, hand-labeled masks, and the accompanying STAC catalog, is archived on Zenodo~\cite{mukherjee_2025_ufo} under a CC BY 4.0 license (DOI: \href{https://doi.org/10.5281/zenodo.19698577}{10.5281/zenodo.19698577}).

\section*{Code availability}
Code for training the deep learning model on UFO and for evaluating predictions against UFO labels is available at \url{https://github.com/Tellman-lab/urbanFloodObservations}.

\bibliography{ufo_ref}

\section*{Acknowledgments}
We would like to thank Simone Holliday, Patrick Hellmann, Natasha Rapp, and Linn Ji for their assistance with labeling. The creation of the UFO dataset was supported by a NASA Terrestrial Hydrology Program grant (\#80NSSC21K1044). This research was partially supported by the U.S. Department of Energy, Office of Science, Biological and Environmental Research (BER) as part of the Integrated Coastal Modeling (ICoM) project (Grant KP1703110/75415).

\section*{Author contributions statement}

B.T., R.M., and H.F. conceived the dataset; H.F. selected flood events and PlanetScope imagery and managed the labeling team; R.M. performed final label quality control and selection, trained the models, and prepared the dataset; A.I. performed initial label quality control; R.M. and Z.Z. prepared the STAC catalog; R.M. and H.F. wrote the manuscript; B.T., Z.Z., and J.G. revised the manuscript; J.G. assisted with data processing; and B.T., U.L., and V.L. supervised the project. All authors reviewed the manuscript.

\section*{Competing interests}

B.T. holds equity in Floodbase, a company that sells near-real-time flood mapping and monitoring systems.

\end{document}